# Deformation measurement of a soil mixing retaining wall using terrestrial laser scanning

Yang Zhao[1], Lei Fan[1,*], Hyungjoon Seo[2]

1. *Department of Civil Engineering, Design School, Xi'an Jiaotong-Liverpool University, 111, Ren'ai Road, Suzhou Industrial Park, Suzhou, China*
2. *Department of Civil Engineering and Industrial Design, School of Engineering, University of Liverpool, Liverpool, L69 3BX, UK*

*Corresponding author: email: Lei.Fan@xjtlu.edu.cn

Retaining walls are often built to prevent excessive lateral movements of the ground surrounding an excavation site. During an excavation, failure of retaining walls could cause catastrophic accidents and hence their lateral deformations are monitored regularly. Laser scanning can rapidly acquire the spatial data of a relatively large area at fine spatial resolutions, which is ideal for monitoring retaining walls' deformations. This paper attempts to apply laser scanning to measurements of the lateral deformations of a soil mixing retaining wall at an ongoing excavation site. Reference measurements by total station and inclinometer were also conducted to verify those from the laser scanning. The deformations derived using laser scanning data were consistent with the reference measurements at the top part of the retaining wall (i.e., mainly the ring beam of the wall). This research also shows that the multi-scale-model-to-model method was the most accurate deformation estimation method on the research data.

## 1. INTRODUCTION

Retaining walls are often constructed before a basement excavation to prevent excessive movements of the natural ground and the adjacent structures. Failure of retaining walls during construction could lead to severe accidents. Deformation is an important index to reflect the structural response of retaining walls to excavations. As such, it is essential to monitor the deformations of a retaining wall routinely during excavations. According to the Technical Code for Monitoring of Building Excavation Engineering [1], such monitoring is also mandatory in China. In practice, there are various means of deformations monitoring of retaining walls, including geotechnical instruments (e.g., inclinometers and tiltmeters) and traditional surveying techniques such as total station, GPS, tacheometry, inclinometer. However, in these methods, deformations are measured only at discrete locations. In addition, some of those could also come with a high labour cost and low efficiency [2-5].

Terrestrial laser scanning (TLS) can rapidly acquire fine resolution measurements on a large area in a non-contact way and does not require any instrumental parts to be installed at the objects measured [6]. These features suggest that TLS is potentially a good means of monitoring the deformations of retaining walls, where the deformations are typically estimated through a comparison between multi-temporal point cloud datasets obtained in laser scans. In the literature, this technique has been widely applied for monitoring the deformations of various types of engineering structures, including retaining walls [6-9], historical architectures [10-13], bridges [14-16], arch structures [17-19], tunnels [3, 20, 21], dams [22], beam elements [23, 24] and hull sections [25].

The feasibility of using TLS to monitor a retaining wall was assessed by Laefer and Lennon [6]. They found that while the technique offered additional benefits, its effectiveness was still doubted. Oskouie, Becerik-Gerber and Soibelman [7] measured vertical settlements of a retaining wall along a highway corridor. In their study, the joints between the concrete facing panels of the wall were extracted, the vertical deformations of which were used to represent the vertical deformations of the retaining wall. Their focus was on the means of extracting the joints. However, an appraisal of the quality of the final deformation results was not conducted, not only because no alternative measurements were available but also because the multi-temporal scans were registered roughly using static geometric features near the retaining wall due to a lack of artificial reference targets at the site. Lin, Habib, Bullock and Prezzi [8] scanned a mechanically stabilised earth retaining wall using TLS. The point cloud data were analysed to obtain the out-of-plane offset and angle deviations of each individual concrete panel on the wall. The out-of-plane offset was the distance between a panel and the best-fitting plane of the wall in the normal direction of the wall. The angle deviations are the difference between the longitudinal directions of the surrounding joints of a panel and orthogonal directions (horizontal or vertical). Those analyses were carried out in a point cloud acquired at a single time epoch. Thus, registration of multi-temporal point clouds was not required. Adamson, Alfaro, Blatz and Bannister [9] measured the vertical settlements of mechanically stabilised earth retaining walls at the abutments of a bridge. The bottom surface of the bridge deck was used as a reference datum of vertical displacement. The results were validated against total station measurements and finite-element-modelling results. However, due to the fact that the bridge deck was horizontal, only vertical movements of the retaining wall could be estimated. Horizontal deformations [26] and tilting [27] of the concrete panels of a highway retaining wall were measured using TLS. The multi-temporal scans were registered using reference targets. Kalenjuk, Lienhart and Rebhan [28] also measured tilting and horizontal deformations of an anchored retaining wall along a highway using a mobile laser scanner mounted on a vehicle. The registration of scanned point clouds was achieved by GNSS positioning information. However, the results of Seo [26] and Kalenjuk, Lienhart and Rebhan [28] were not verified against alternative measurement methods.

For deformations estimated using point cloud data, it is helpful to understand the likely level of uncertainty. It is typically a combination of three parts, including the instrumental error of the scanner used, the registration (or georeferencing) error, and the modelling error [29]. A brief introduction to these components is given as follows.

- The instrumental error is the difference between the measured position of a point and its true position. The quantification of the instrumental error requires an investigation of the mechanical, electronic and optical sources of the error for calibration and correction [30]. In addition, it is also affected by factors such as incidence angle, scan distance, atmospheric conditions, and surface characteristics of the observed objects [6, 29, 31-39]. Luckily, most TLS scanners available (such as the one used in this study) have a very high measurement accuracy.

- Registration is a fundamental task for the comparison between multi-temporal scans. Achieving a high-quality registration is one of the main challenges to laser scanning surveys at a dynamic construction site. Registration using reference targets is by far the most commonly used registration method in practice. If implemented appropriately, target-based registration is more accurate than those based on feature points and surface-matching [40, 41]. Fan, Smethurst, Atkinson and Powrie [40] proposed empirical equations to estimate the expected registration error based on numerical simulation and discussed various influencing factors that need to be taken into account when the layout of the targets is arranged at sites.
- As point cloud data are unstructured, there will be no direct point-to-point correspondences between multi-temporal point cloud data. As such, modelling is often required to enable a comparison between two point clouds for deriving surface deformations. The error result from the difference between the model and the original point cloud are termed modelling error [29]. Researchers [29, 32, 40] already recognised that the modelling error depends on the point cloud density, the complexity of the object surface scanned, and the modelling method adopted. In some deformation estimation methods, such as Cloud to Cloud(C2C) method, direct point-to-point estimation is enabled without resorting to modelling. So, when referring to the error result from the deformation estimation process, the term 'deformation estimation error' is used instead of modelling error hereafter. The magnitude of the deformation estimation error is case-specific depending on the characteristic of the point cloud itself.

In practice, it is impossible to quantify the gross error of each data point in a point cloud because their ground truth is unknown, and it is challenging to obtain benchmark data of much higher quality than those from laser scanning. However, it might be helpful to carry out statistical studies using resampled data to understand the relative level of uncertainty in the estimated deformations.

The application cases of TLS on the monitoring of retaining walls are still rare in the literature. In those reported [7-9, 26-28], few used reference measurements to validate the deformations measured by TLS. As such, its effectiveness could not be evaluated. In addition, the surfaces (e.g., steel or concrete surface materials) of the retaining walls monitored in those cases were comparatively smooth, which does not represent all types of retaining walls. To the authors' best knowledge, no research was found to monitor the deformations of soil mixing retaining walls at an ongoing excavation site, which have a rough and soft wall surface. The uncertainty involved in applying TLS in such a scenario is not well known and thus was explored in this study.

## 2. MATERIALS AND METHODS

### 2.1. Study site

The construction site considered is a part of the basement excavated for a new warehouse building. Before the excavation, a Soil Mixing Wall (SMW) was constructed along the basement perimeter to be built. The depth of the SMW ranged from 12-21 meters. A reinforced concrete ring beam connected the crest of the SMW. There are two excavation depths: 10.3 m in Section A to accommodate a two-storey basement and 5.3 m in Section B for a one-storey basement, as shown in FIGURE 1. For relatively deep excavations (deeper than 10 m), such as in Section A, lateral props are typically built to support the retaining walls before the excavations are carried out. In this case, the lateral deformations are likely to be comparatively small. In addition, the presence of the lateral propping system can lead to a large number of occlusions when a terrestrial laser scanner is positioned at the ground level (entering the area inside the exaction was not permitted). Only the retaining wall in Section B was selected for laser scanning surveys for those reasons. The laser scanner could only observe the excavated surface of the retaining structures. Note that the excavation depth (5.3 m) was less than half the total depth (12 m) of SMW in Section B. Whether the monitored portion of the SMW can sufficiently represent the deformation map of SMW over its entire depth should be considered.

The excavation started in Section A on 13$^{th}$ March 2018 and was completed on 14$^{th}$ April. From 17$^{th}$ May to 25$^{th}$ May, Section B was excavated to 4 m below ground level. On 25$^{th}$ May 2018, the local authority suspended the excavation in Section B due to control of dust production activities. On 25$^{th}$ July, the excavation in Section B resumed and finally reached the design depth of 5.3 m on 28$^{th}$ July. For ease of discussion, the excavations before and after the suspension are referred to as excavation stage 1 and stage 2, respectively. As required by the excavation protocol, a certain thickness of soil should be left on the surface of retaining structures during excavation. In order to avoid damage of retaining structures by excavation works. Loose soil remaining on the vertical surface of retaining structures are susceptible to erosion. The laser scanning measures the surface of retaining structures. The deformation map based on point cloud analysis will also include the effect of superficial change induced by the loss of loose soil on the retaining structures.

The deformations of the SMW and the ring beam connecting the SMW at its crest were measured. The Technical Code for Monitoring of Building Excavation Engineering [1] specifies the alerting thresholds for the deformation monitoring of ring beams (3 mm daily and 25 mm cumulative) and SMW (4 mm daily and 50 mm cumulative).

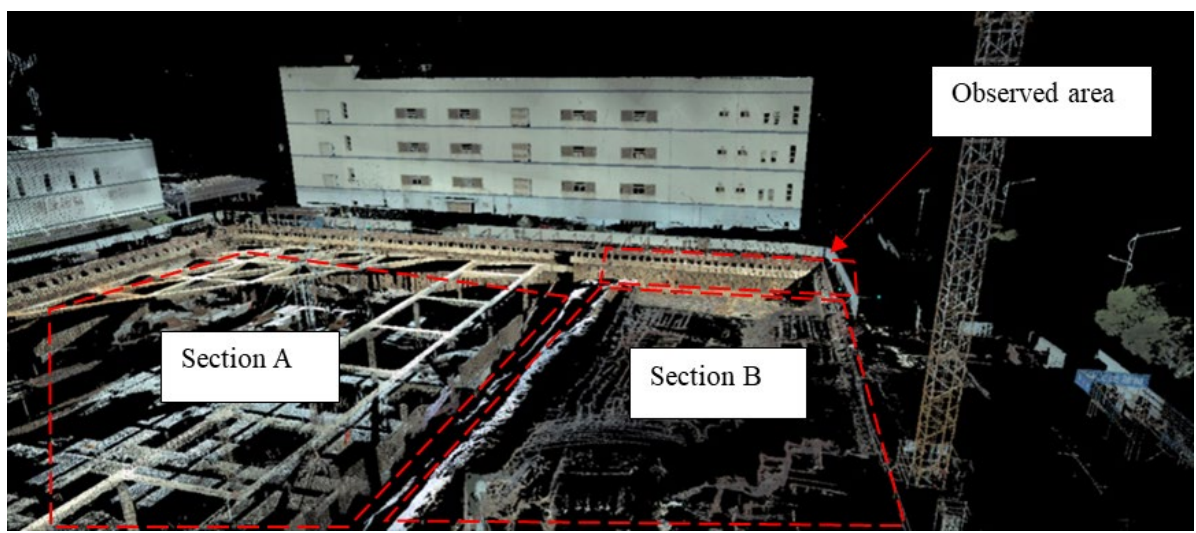

FIGURE 1: Site layout.

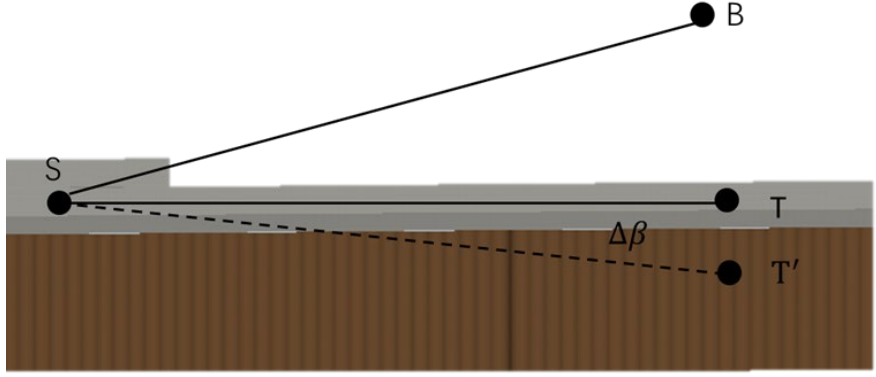

FIGURE 2: Principle of small angle approximation method.

## 2.2. Reference data

A total station (RTS010, FOIF) was used for the measurement of ring beam deformation using the small angle approximation method. Its angular measurement accuracy was 1″ and its distance measurement accuracy was 2 mm + 1 ppm (parts per million). Steel rods were cast in the ring beam, serving as the monitoring target. As shown in FIGURE 2, the total station was positioned in such a way that the line of sight connecting the total station (*S*) and a monitored target (*T*) is approximately parallel to the longitudinal direction of the ring beam (see the layout in FIGURE 2).

*B* represented the tip of a lightning rod on the top of a factory building approximately 80 metres away, which was used as the benchmark and assumed stationary. *T′* was the temporal location of *T* after some movement. $\beta$, representing the initial value of angle $\angle BST$ was measured by total station and was recorded as a baseline. The angular difference $\Delta\beta$ between $\angle BST$ and $\angle BST'$ was used to estimate the deformation *D* from *T* to *T′* using Equation [1]. Equation [1] is an approximation based on the assumption that $\beta$ is small and hence arc and chord length from *T* to *T′* are equivalent.

$$D = \frac{\Delta\beta}{\rho} L \qquad [1]$$

where $\Delta\beta$ is the angular difference between $\angle BST$ and $\angle BST'$ in seconds; $\rho = 360°\times180°/\pi$, which is a unit conversion factor; *L* is the length of line ST in metres.

An inclinometer (GHHB-CX-30-B; Gemho) was used to measure the lateral deformations of the ground over the depth. The schematic drawing of the inclinometer measurement is shown in FIGURE 3. It may be assumed that the deformation measured by the inclinometers represent the lateral deformations of the retaining wall over its depth. For readings, the inclinometer probe was placed down to the bottom of the tube pre-installed in the ground and then lifted up every half metre. During the gradual lifting process, the probe records the tilting angle $\theta$ every 0.5 metres of travel. The horizontal deformation within the 0.5 m intervals can be estimated as $0.5 \times sin\theta$ m. It is assumed that the bottom of the tube is stationary, and the lateral deformation at the bottom is therefore zero. The deformation map over depth was obtained by accumulating the incremental deformations of each 0.5 meter upwards. The deformation map over depth was also taken daily. The angular accuracy of the inclinometer is 0.01°, based on which the accuracy of lateral deformation measurement over a unit depth is $1 \times \sin(0.01°) = 0.00017$, i.e., 0.17 mm per m.

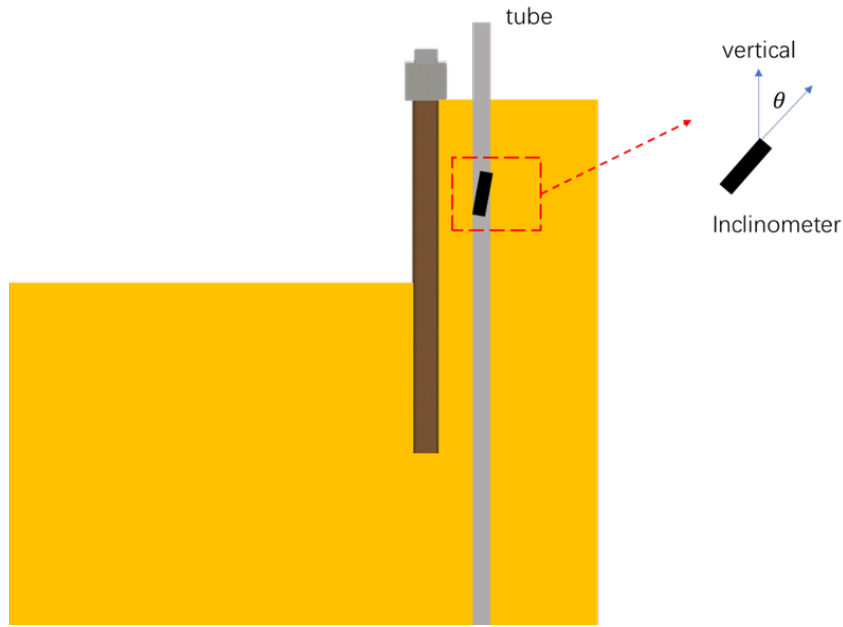

FIGURE 3: The side view of the inclinometer and the retaining structure.

## 2.3. Point cloud data

To enable a comparison between the multi-temporal point clouds, one of those should be used as the reference. The others need to be transformed into the same coordinate system as the reference. Therefore, it requires careful consideration of a suitable registration strategy before any survey campaigns are carried out. Using dedicated artificial reference targets effectively achieves high registration accuracy in many cases if this approach is implemented appropriately at sites. However, it was found challenging to identify suitable objects to accommodate the targets for the construction site considered in this study. This is because limited places within the construction site were granted access to for setting up the targets and because the objects in reasonable distances to the area being scanned are likely to move over time. After taking into account various trade-offs, a set of black and white paper targets were arranged on the fencing panels used to isolate the construction site from the public space. Each fencing panel was fixed in position at its bottom and was laterally restrained by an inclined prop. However, their actual stability conditions were not known. It was assumed during the data collection stage that the target-based approach might not be a reliable solution for the site. As a backup solution, the facade of a factory building adjacent to the construction site was also recorded by the scanner intentionally using a very fine scanning resolution. It has features (such as regularly spaced windows and joints) that could enable a high-quality cloud-to-cloud registration. In addition, as part of the site monitoring, the vertical settlement of the building façade was also measured daily to check that the deformations under the effect of the ongoing excavation were within the limit required by the building owner. These measurements were considered for likely correction in the cloud-to-cloud registration.

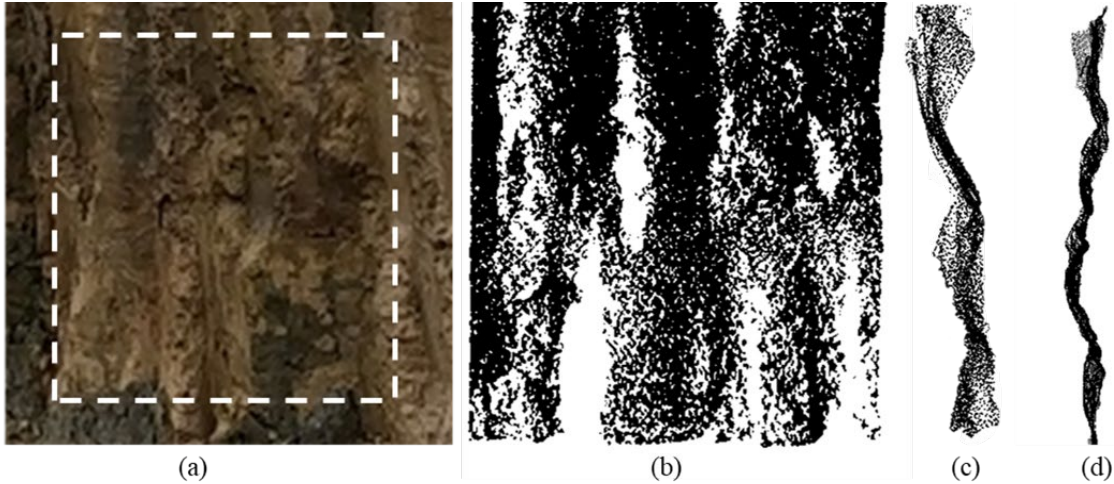

FIGURE 4: The SMW wall surface: (a) a photo showing the surface complexity; (b) the front view of the point cloud corresponding to the area highlighted by the dashed rectangle in (a); (c) the side view of the point cloud in (b); (d) the top view of the point cloud in (b).

The surface of the SMW is very rough and displays a wavy pattern (as shown in FIGURE 4), which means that local occlusions exist in the point cloud data acquired from a single scanner location. The SMW area of interest needs to be scanned from multiple locations to form registered point cloud data to minimise the data occlusion. However, this was not possible given the constraints of the construction site considered, as suggested by the site engineers. On the Northern, Eastern, and Western sides of the site, existing factories are in service. On the Southern side of the site, there was a public road in service. The fencing wall is two metres tall and blocks the view of the site from the outside. Thus, the scanner could be placed only within the construction site. It was also prohibited to set the scanner inside the excavated ground. The only available candidate locations were the narrow strip area between the excavation and the perimeter wall. Also, the location of the scanner could not interrupt the frequent travel route of heavy vehicles and construction machines. With all these constraints, the area permitted for setting up the scanner is rather small. It was observed that arranging the scanner at different locations within the area permitted did not effectively change the angles of view from the scanner to the retaining wall scanned. Therefore, as a compromise, the wall was scanned from only a single chosen location, which would inevitably have led to local data gaps in the point cloud obtained.

A laser scanner (Leica ScanStation P40) was used onsite. Its 3D position accuracy is 3 mm at 50 metres. The tilting angle of the scanner was limited to 2 seconds, and the spatial resolution was set as 2 mm at a 10 m distance. Laser scanning campaigns were conducted on 29$^{th}$ June and 23$^{rd}$ August 2018.

### 2.4. Methods for calculating deformations

Using the registered multi-temporal point cloud data, the deformations of the retaining structure can be estimated. The deformation estimation is conducted in a local coordinate system, which was defined as follows,

- The z-axis is in the vertical direction, and upwards is positive.
- The y-axis direction is perpendicular to the retaining structures, and the negative direction is towards the excavation pit.

There are various types of methods for this task. However, because the performance of each method varies with the surface characteristics of point cloud data, it was difficult to choose the optimal method for the datasets considered in this study. As such, four different methods were adopted, namely Cloud to Mesh(C2M), Mesh to Mesh(M2M), Multi-Scale-to-Model(M3C2) [29] and point-to-plane Iterative Closest Point (ICP) [42]. For ease of discussion, one of the two point clouds compared is named "reference point cloud" (the one obtained in June) and the other "query point cloud" (the one obtained in August). The four methods considered are introduced briefly in the following.

In C2M, a Triangular-Irregular-Network (TIN) mesh (based on Delaunay's triangulation) is generated using the reference point cloud, followed by the calculation of the distances from individual data points to the mesh. The deformation values can be positive or negative, depending on the relative location of local query data points to the mesh.

In M2M, TIN meshes are generated for both the reference and the query point clouds. A regular grid of interpolation locations was specified on a plane parallel to the best fitting plane of either the reference or the query point cloud. TIN meshes are fitted to reference and query point cloud, respectively, using Delaunay's method. The distance between the meshes of the reference point cloud and query point cloud in the y axis-direction is regarded as the M2M deformation. In this research, the size of the grid was set to be 20 mm.

M3C2 is local-normal-based distance estimation method [29]. It calculates the distance between the query and the reference point clouds in the direction of local normal. Firstly, core data points can be sampled from the reference or the query point cloud according to a user-defined core point resolution, which is the minimum distance between any two core points. Secondly, for any given core point $i$, its surface normal is estimated by fitting a plane to data points of the original reference point cloud, from which the core points are subsampled, within the $D_n/2$ distance of core point $i$. $D_n$ is the diameter of the sphere for normal estimation. Thirdly, a cylinder of radius $D_d/2$ is defined so that its axis is in the direction of the estimated normal and goes through the core point $i$. The height of the cylinder can be infinite or a user-defined value $h$. All data points (in the reference and query point cloud) encompassed by the cylinder are projected to the axis. Two mean positions (one for the reference point cloud and the other for the query point cloud) are obtained. Finally, the distance between those two mean positions is computed and regarded as deformation using the M3C2 method. In this research, all points in the reference point cloud are considered core points without subsampling to achieve a maximum resolution of the deformation map. $D_n$ and $D_d$ is set to be 0.03m, which is approximately four times of mean data-spacing. The value of $h$ is 4 m, which is practically infinite considering the expected deformations are in centimetres.

In the ICP-based method, the query point cloud is registered to the reference point cloud using point-to-plan ICP. by optimising the objective function used in the point-to-plane ICP method, a corresponding point in the reference point cloud is identified for each point in the query point cloud. In optimisation of the objective function of point-to-plane ICP, data point H-beam on top of the ring beam was utilised to enhance the point-to-point correspondence. The distance in the y-axis direction between the corresponding point pairs can be estimated and is regarded as deformation using the ICP-based method.

## 2.5. Comparison of deformation estimation error

The performances of the aforementioned C2M, M2M, M3C2, and ICP-based methods depend on the spatial characteristics of the point cloud data and need to be evaluated for the case considered. To evaluate the performance of each method, the estimated values of deformations need to be compared with the true values, which are unknown. Therefore, statistical resampling of the same original point clouds [34, 43] or synthetically generated point clouds [7, 29, 40] can be considered to test the minimum level of detection of each deformation estimation method. In this approach, the original point cloud dataset is subsampled as uniformly as possible into two subsets where they do not share the same data points. These two subsets are used as the reference and query point clouds, respectively, for determining the minimum level of detection that is free of any registration or georeferencing uncertainties. Since the reference and the query point clouds come from the same dataset, the theoretical values of deformations should be zero. The deviations of the deformation values estimated from zero can be deemed to be caused by a combination of the modelling error and the instrumental measurement error.

The error of the deformation estimation method was tested on the point clouds representing SMW and ring beam extracted from real-life scans. Each point cloud was divided into two subsamples uniformly. Before the deformation estimation, the point cloud samples are levelled in the x-y plane, and a minimum bounding box is built for each point cloud. The data spacing, $S$, is estimated following Fan and Atkinson [43] using Equation [2].

$$S = \frac{\sqrt{L_x \cdot L_y}}{\sqrt{N} - 1} \qquad [2]$$

where $L_x$ and $L_y$ are the extents of the bounding box in the x and y-axis, respectively, $N$ is the total number of points.

Researchers [29, 32] already recognised that the deformation estimation error of point cloud deformation estimation is dependent on the point cloud data-spacing. Therefore, it is necessary to establish the minimum detection level concerning variable data-spacing. To achieve that, the query point cloud and reference point cloud are incrementally subsampled using a random-in-voxel method. As we still want the points in these incremental subsamples roughly evenly distributed, voxel subsamples will be appropriate for this matter. Nevertheless, the voxel subsample will take the average position of all points in a voxel, which is contradictory to keeping the points existing in original point clouds. So, a random point in each voxel is selected instead of taking the average position of all points in each voxel. The voxel size is incrementally increased at a step value of two times the initial data-spacing.

The deformation between each pair of subsampled point clouds labelled as a reference point cloud and query point cloud, respectively, was estimated using C2M, M2M, M3C2, and ICP-based methods at incrementally increased data-spacings. The parameter $D_n$ and $D_d$ for M3C2 were adaptively set to 4 times of the current data-spacing. The Mean Absolute Error (MAE) can be determined with the assumption that the ground truth is 0. The MAE was used to represent the level of error instead of mean error because the mean error could be misleadingly low due to the neutralisation of positive and negative values of error [32]. Afterwards, the data-spacing of query point cloud and reference cloud is incrementally increased to study the variation of deformation estimation error with the density of the point cloud. The minimum level of detection of deformation estimation between point clouds of variable data-spacing can be determined.

## 3. RESULTS

### 3.1. Registration results

The targets mounted at the site were used to register the multi-temporal point cloud datasets. However, the registration results show clear misalignment of some comparatively stable objects (e.g., the adjacent building), likely because the fencing where the targets were installed moved between the two scans. Such results were not surprising due to the possible issues with the setup of the artificial reference targets, as stated in Section 2.3. As such, the backup registration solution using the facade of the nearby building was adopted for a Point-to-Plane ICP-based registration. The building facade was monitored regularly and found to have negligible deformations.

The settlements of 6 control points (as shown by red points in FIGURE 5 of the building façade between the two TLS campaigns were measured using a total station (RTS010, FOIF). At these checkpoints, the magnitude of the settlements ranged from 1.46 mm to 3.01 mm (as shown in FIGURE 6), which was assumed to be negligible. The lateral deformation data were not made available, but the site engineer suggested that they were smaller than the settlements. As such, the building façade can be used as a good reference object for the registration.

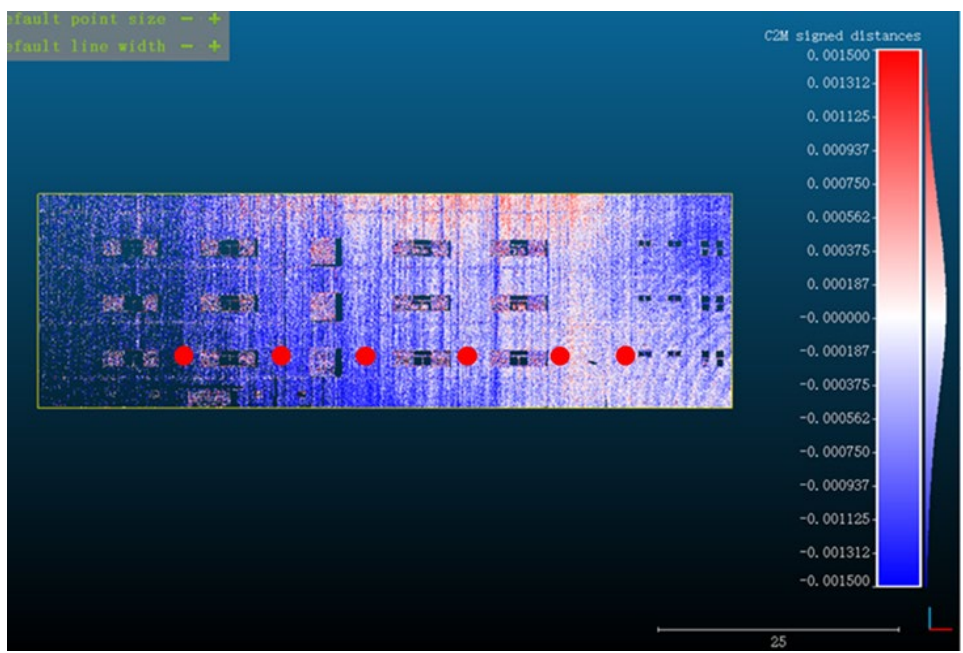

FIGURE 5: Differences (estimated using C2M) between the point clouds (representing the building facade) obtained in June and August.

The point-to-plane ICP was conducted, in which the RMSE of inlier correspondences was 0.05 mm. The distances between the two registered point clouds were estimated using the C2M method to check the goodness of the registration. The distance map is shown in FIGURE 5. The maximum absolute value is 1.65 mm, and the mean value is 0.005 mm. In addition, the spatial distribution of those values is trendless, which indicates that the differences were likely to be caused by the random measurement noise in the individual data points.

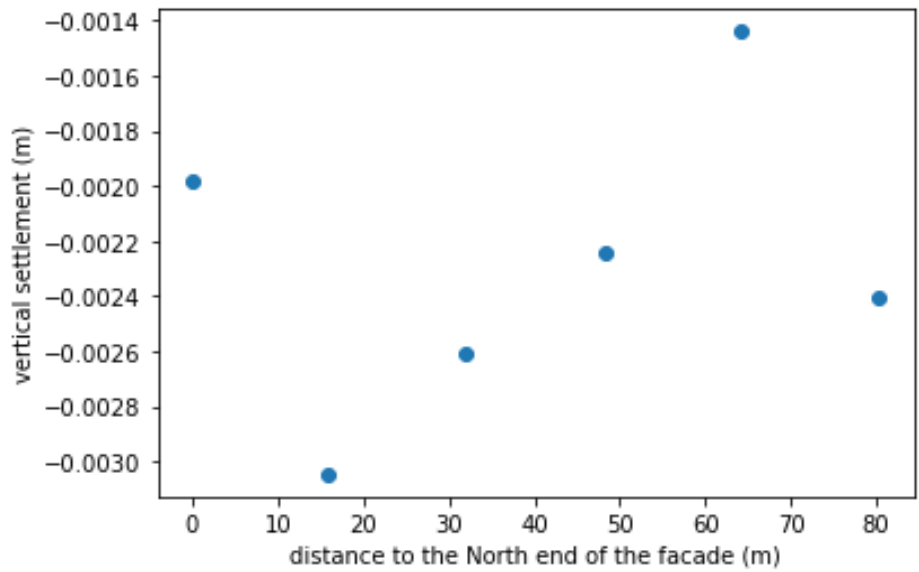

FIGURE 6: The vertical settlements of the building façade between the scans in June and August.

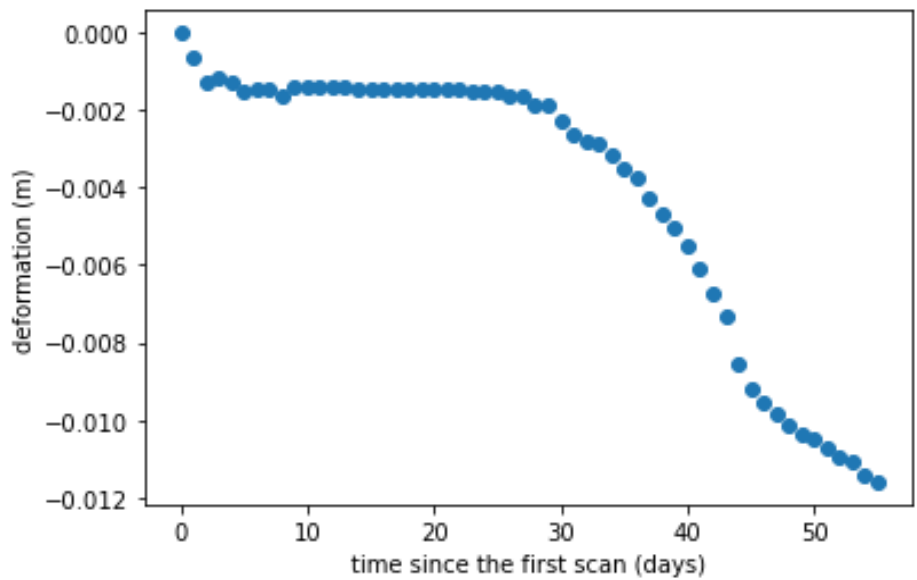

FIGURE 7: The deformations of the pile crest since the first scan date.

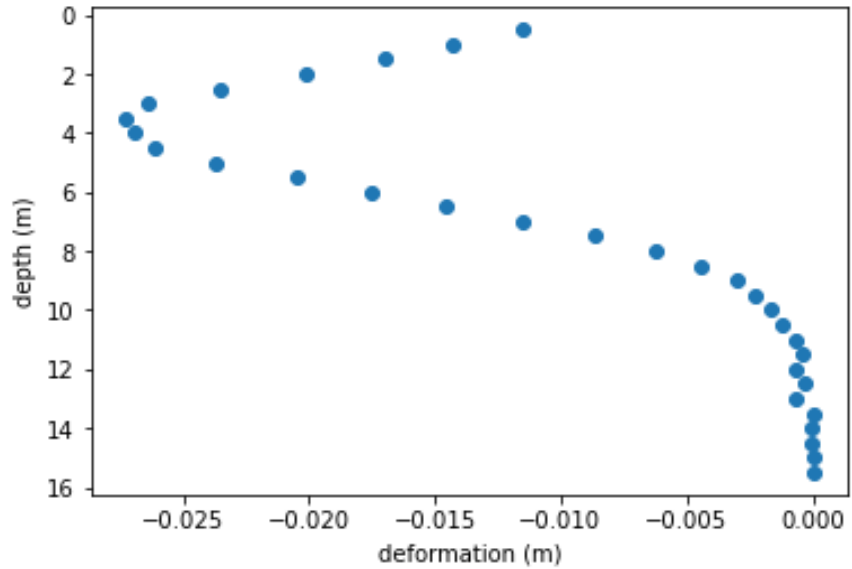

FIGURE 8: The lateral deformations of the retaining wall over depth from June to August.

## 3.2. Benchmark deformations

### 3.2.1. Deformations of the ring beam measured by the total station

As shown in FIGURE 7, in the first six days from the initial scan, the pile head moved approximately 1.8 mm towards the excavated side due to excavation in stage one. From day 7 to day 30, the deformations were less than 0.4 mm, suggesting that no actual deformation occurred. The measurement error likely caused the variation in the measured values. This also indicates that the total station measurements had relatively small errors. On day 30, the excavation resumed and caused further deformations. The deformation increased by approximately 10 mm from day 30 to day 58. The total deformation from day 0 to day 58 was approximately 11.6 mm, in each of which a laser scan was carried out.

### 3.2.2. Lateral deformations over depth by the inclinometers

FIGURE 8 shows the lateral deformations of the retaining wall over depth, which were measured by the inclinometer during the period between the initial scan and the subsequent scan considered. The deformation monotonously increased with the increase of depth from top to 4 m of depth. The deformation on the top (depth 0.5 m) was 12 mm and reached its maximum value of 27 mm at a depth of 4 m, which was 1.3 m above the maximum excavation depth. The deformation decreased monotonously towards the bottom of the inclinometer tube at a depth of 22.5 m.

## 3.3. Deformations derived using point cloud data

FIGURE 9(a) to (d) show the deformation maps obtained by applying C2M, M2M, M3C2, and ICP, respectively, to the point cloud data acquired in June and August. No clear patterns of deformations can be observed from the deformation maps as the scale of the colour bars is stretched by extreme values. For example, the extreme positive values (which indicates the direction of deformation is away from the excavation pit), which were likely caused by the loss of loose soil on the SMW surface, could be as much as. 0.4 m. On the other hand, the extreme negative values (which indicates the direction of deformation is into the excavation pit) were likely caused by soil deposition around the intersection areas between the SMW and the excavated ground surface. Those extreme values due to loss and deposition of loose soils are probably caused by unintentional human disturbance during the subsequent construction activities after excavation.

Thus, according to the range of the deformations measured by the inclinometer, extreme values were filtered out. Only those within the range from -15 mm to 0 were shown in FIGURE 10. After the filtering, the patterns in the deformation maps are visible and consistent between the approaches (apart from the ICP method where deviations from the others were observed locally) used to derive those maps. The spatial variation of deformation values in the map derived using the ICP-based method is larger than those from other methods because the distance calculation using ICP is based on point-to-point correspondence, which is subject to the range noise of individual points.

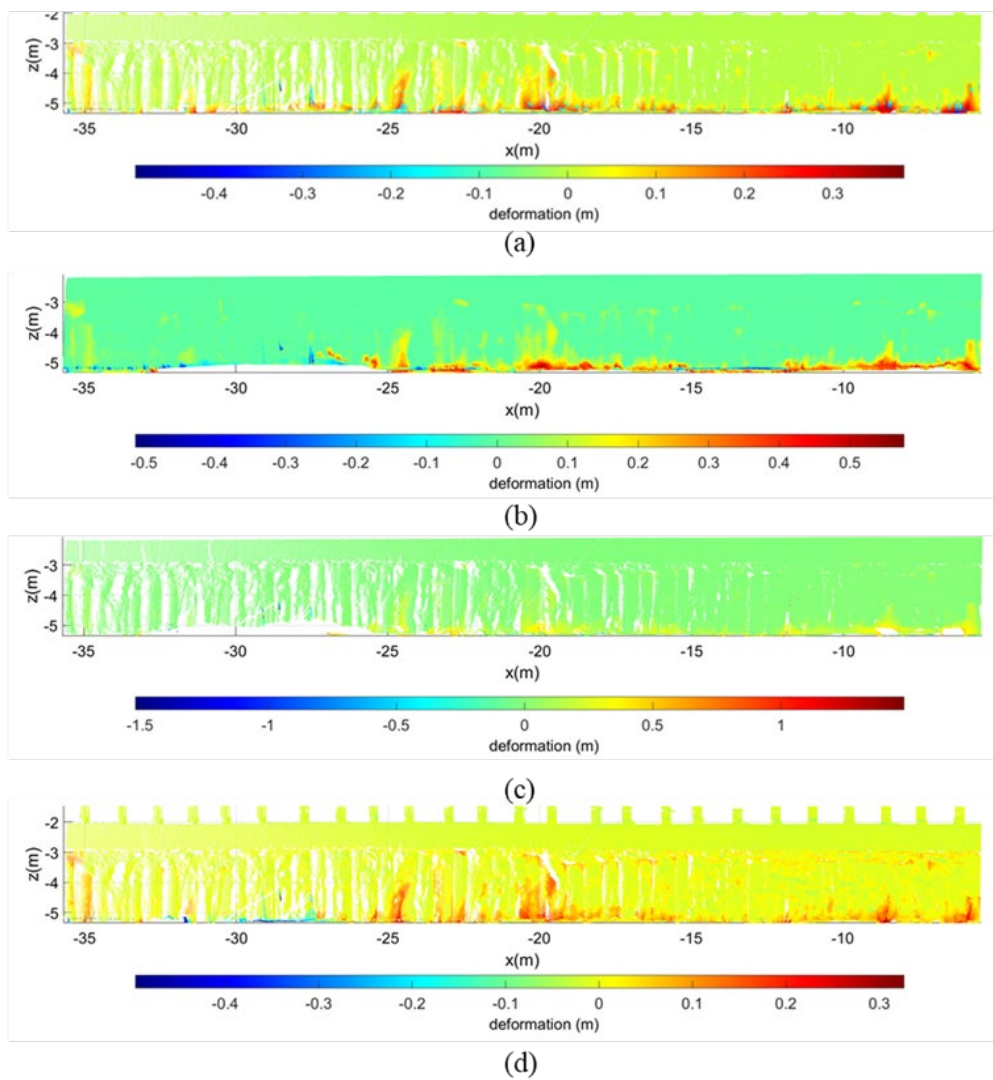

FIGURE 9: The deformation maps of the retaining wall using: (a) C2M; (b) M2M; (c) M3C2; (d) ICP.

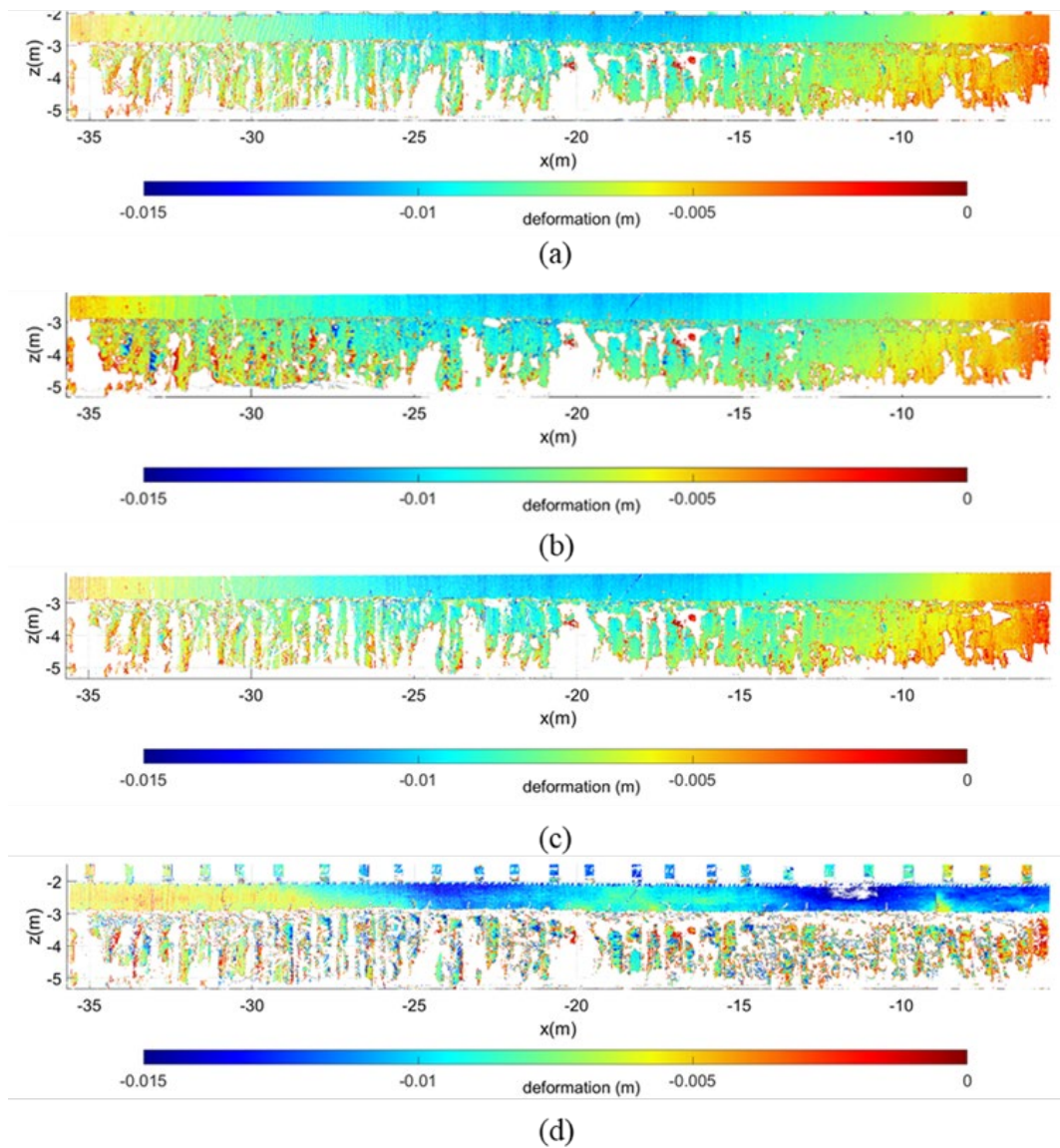

FIGURE 10: The filtered deformation maps of the retaining wall using: (a) C2M; (b) M2M; (c) M3C2; (d) ICP.

Along the length of the retaining wall section considered, the deformations (obtained using C2M, M2M, and M3C2) were observed to be largest in the middle, moderate on the left, and smallest on the right. These observations are expected and match the propping conditions in the field. The retaining wall section considered does not have any transverse support, in which case the transverse deformation at the mid-span is expected to be the largest. However, beyond its left side, there is another and deeper excavation (i.e., Section A in FIGURE 1) where the retaining wall was supported by bracing of beams. The right side of the section considered was supported by an orthogonal retaining wall along with its whole depth, so the overall stiffness of this support is larger than that of the lateral propping at the left. Over the depth of the wall, the deformations increased slightly, but the rate of change is not as large as that seen in the inclinometer measurements.

The deformation in the middle of the ring beam is 11.8 mm according to total station measurement and 11 mm according to point cloud analysis. The difference is 0.8 mm.

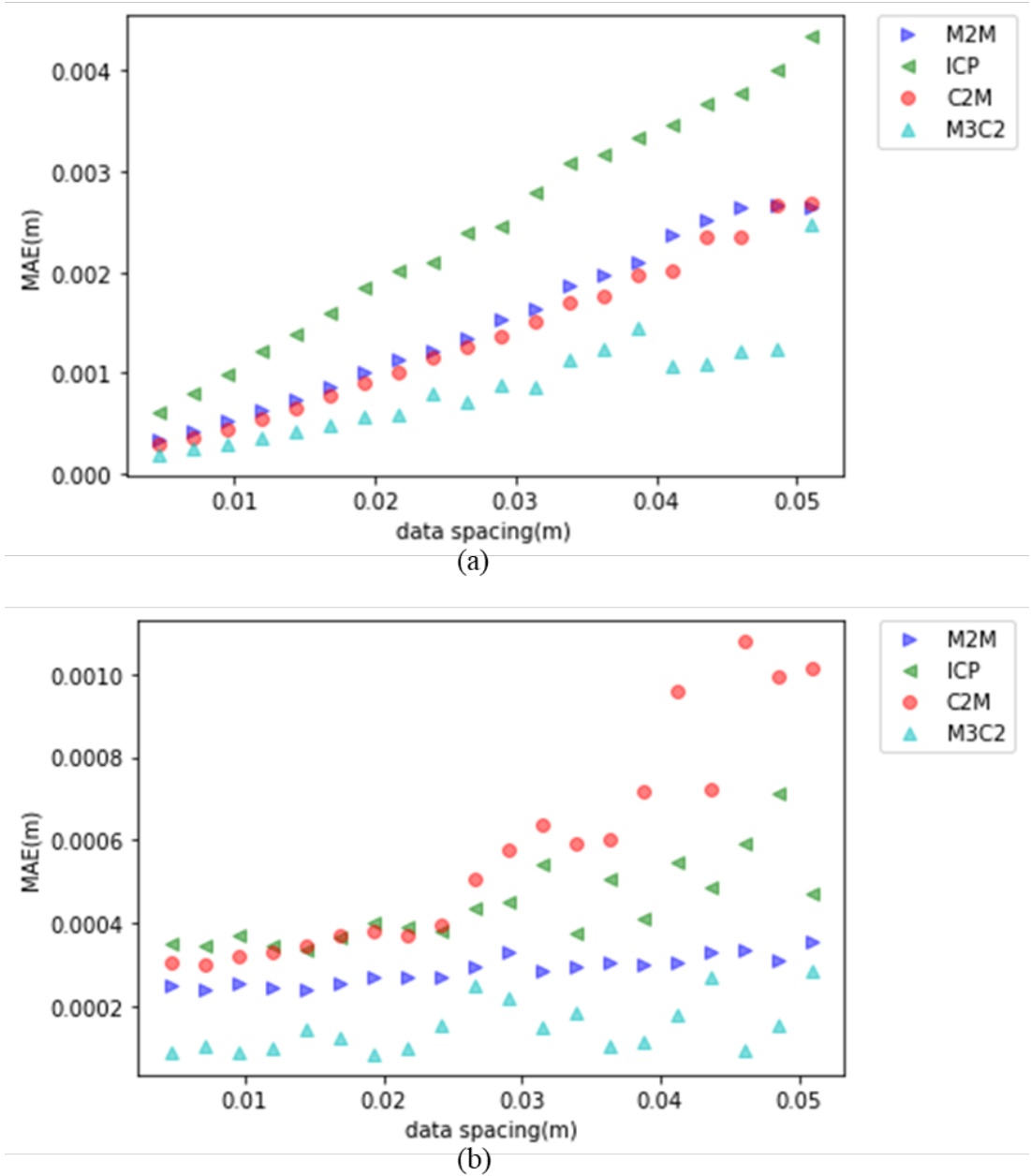

FIGURE 11: The MAE in the deformations estimated using various methods: (a) the SMW; (b) the ring beam.

### 3.4. The MAE of deformation estimation

FIGURE 11 shows the MAE in the deformations estimated at different levels of data point density (represented using the data spacing) using the four methods (ICP, C2M, M2M, and M3C2) for the point cloud data representing the SMW (FIGURE 11(a)) and the ring beam (FIGURE 11(b)). For the point clouds of both cases considered, the MAE increased with increasing data spacing. MAE in M3C2 was the lowest. In the M3C2 method, the estimation of deformations is based on the average position of multiple points in a neighbourhood instead of the position of a single point. This mechanism mitigates the effect of the range noise in individual points.

Similarly, by building a TIN in the C2M or M2M methods, the deformation estimation would not rely on point-to-point correspondences. The MAE in the ICP method was the largest because the point-to-point correspondences might be far-fetched due to the roughness and data occlusion. The data spacing of the retaining wall point cloud varied with the distance to the laser scanner but were all smaller than 0.01 m. At the data-spacing of 0.01 m, the minimum level of detection of the ring beam part of the retaining wall ranged from 0.05 mm to 0.4 mm for the four different deformation estimation methods. At the data-spacing of 0.01 m, the minimum level of detection of the SMW part of the retaining wall ranged from 0.2 mm to 1 mm for the four different deformation estimation methods.

## 4. DISCUSSION

The minimum level of detection without the effects of registration error in the numerical simulations is presented in Section 3.4. As expected, the minimum level of detection generally increased with data-spacing. The M3C2 method had the least amount of the minimum level of detection.

The registration results are presented in Section 3.1. The small amount of the residual mean difference between the two scans of the building facade after registration and the small amount of the minimum level of detection in the numerical experiments jointly demonstrate that the deformation maps in FIGURE 9 and FIGURE 10 should be adequately accurate.

The deformation maps in FIGURE 9 and FIGURE 10 show that in the middle of the ring beam part of the retaining wall, the deformation value is consistent with that measured by the total station (see FIGURE 7). In similar previous studies [7-9, 26-28], such validation was not provided except in the study of Adamson, Alfaro, Blatz and Bannister [9]. Horizontally, the deformation was the largest in the middle of the retaining wall and gradually decreased to both ends. This is because, at the two ends, the retaining wall was transversely supported. Vertically, the deformation values were similar from the top to the bottom of the retaining wall in the deformation maps. Such a pattern is different from the vertical deformation profile obtained using an inclinometer (see FIGURE 8). The differences in the deformation values between the laser scanning and inclinometer measurements can be as many as 19 mm at the bottom of the SMW. It is suspected that the differences resulted from the loss of loose soils on the SMW surface. 19 mm thickness of soil erosion is possible during the approximately two-month observation period in a rainy season according to measured natural soil erosion values [44].

Practical concerns experienced in this study should be also be noted. To facilitate the comparison between those multi-temporal datasets, registration and georeferencing are essential steps where all the datasets are transformed into the same coordinate system [22]. To this end, reference targets should typically be installed at positions that are free of any movements over time. However, it is often difficult to identify and verify such locations at ongoing construction sites. The reference targets should be left at the site during the monitoring period or accurately replaced at precisely the same positions. Both options might be complicated for practical reasons. Even if the locations of a laser scanner and the reference targets were cautiously chosen, the line of sight between the reference targets and the scanner might be lost due to dynamics on the site during the monitoring period. Likewise, the line of sight between the scanner and part of the area being scanned. Changing the location of reference targets would potentially cut off the timeline in comparisons between multi-temporal scans. Probably for these constraints, very limited applications of TLS in monitoring retaining walls have been conducted till now.

In this study, the coverage of the measured retaining wall was limited by the availability of space for setting up the scanner. Consequently, there were many local data occlusions in the data collected due to the rough SMW surface. However, if the SMW was scanned from multiple suitable scanner locations, the data occlusions could significantly be reduced. This would lead to denser point cloud data of a more even point distribution over the whole SMW area scanned. Using such datasets, the lateral deformation maps are expected to have higher accuracies.

It is interesting to observe in FIGURE 11(a) that the MAE increased almost linearly with the data spacing. To understand this, it is important to appreciate that the overall errors can be decomposed into two components, consisting of propagated measurement errors and modelling errors (the modelling error reflects how well a surface modelling method approximates the real physiography). Theoretically, the effect of the measurement noise should not change with data spacing [34]. As such, the linear behaviour in FIGURE 11(a) was likely due to the error in the surface model (i.e. the modelling error), which often increased linearly with the data spacing within a certain range, as demonstrated in some previous studies [34, 45].

There are studies (e.g., [46]) on the layout or topology of scanner stations to achieve complete coverage of an area to be scanned. However, it was found that it was impractical to implement a sound survey topology once the construction activity had started. As such, it is recommended that the laser scanning survey arrangement be considered during the planning stage of the construction, which might reserve some spaces for survey campaigns during the construction stage.

## 5. CONCLUSIONS

This research explored the applicability of laser scanning on monitoring the deformations of a soil mixing retaining wall at an ongoing excavation site. The deformations derived using the laser scanning data were checked by the measurements of a total station and inclinometers at some locations. Various uncertainties associated with the application under consideration were reported and discussed. The following conclusion can be drawn.

- The deformations derived using the laser scanning data were similar to those obtained by the total station and inclinometer measurement at the top part of the retaining wall.
- The pattern of the deformation maps along the length of the wall is consistent with the lateral supporting conditions of the wall during construction. However, their pattern along the wall depth is different from the inclinometer measurements. The difference is probably the result of the erosion of the loose soils on the SMW surface.
- This study confirms that it is challenging to implement target-based registration in an ongoing dynamic excavation site, which could be replaced by a registration strategy based on stable buildings adjacent to the construction site. Such stable buildings in an urban environment are typically available because the building owners often require minimum deformations. The results in this study suggest that this alternative can be very effective.

- Amongst the several deformation calculation methods considered, the minimum level of detection derived by the M3C2 method is the smallest for both the ring beam and the SMW of the retaining wall.
- The study shows that the registration error and the minimum detection level could be relatively small, suggesting that laser scanning is an accurate tool for deriving deformations of retaining walls of solid surfaces (e.g., the concrete ring beam). However, its application to the SMW can be significantly affected by the changes (e.g., erosion in a wet season) in the soft soils of the wall surfaces.

**ACKNOWLEDGEMENTS**

The authors are grateful for the financial support from Xi'an Jiaotong-Liverpool University (XJTLU) for the research work, under XJTLU Key Program Special Funding (grant number KSF-E-40) and XJTLU Research Enhancement Funding (grant number REF-21-01-003).

**NOMENCLATURE**

| | |
|---|---|
| $P$ | Unit conversion factor in small-angle method (angular seconds) |
| $D$ | Deformation of the observed target in small-angle approximation method (metres) |
| $L$ | Distance between the observing device and the observed target in small-angle method (metres) |
| $D_n$ | The diameter of the sphere for normal estimation in the M3C2 method (metres) |
| $D_d$ | Diameter of the cylinder to calculate mean position in m3c3 method (metres) |
| $S$ | Point cloud data-spacing (metres) |
| $L_x$ | Extents of the bounding box in the x-axis, respectively (metres) |
| $L_y$ | Extents of the bounding box in the y-axis, respectively (metres) |
| N | Total number of points in a point cloud |

**Greek symbols**

| | |
|---|---|
| $\beta$ | The angle between the baseline and the current sight line in total station measurement (angular seconds) |
| $\theta$ | Tilting angle of inclinometer (angular degrees) |